# Realistic Rendering of Kinetostatic Indices of Mechanisms

CHABLAT Damien, BENNIS Fouad

Institut de Recherche en Communications et Cybernétique de Nantes
1, rue de la Noë
44321 Nantes France
Téléphone 02 40 37 69 54 /Fax 02 40 37 69 54
E-mail: {Damien.Chablat,Fouad.Bennis}@ irccyn.ec-nantes.fr

**Abstract:** The work presented in this paper is related to the use of a haptic device in an environment of robotic simulation. Such device introduces a new approach to feel and to understand the boundaries of the workspace of mechanisms as well as its kinetostatic properties. Indeed, these concepts are abstract and thus often difficult to understand for the end-users. To catch his attention, we propose to amplify the problems of the mechanisms in order to help him to take the good decisions.

## 1- Introduction

This article presents a new approach for the interpretation and the analysis of boundaries of the workspace and the kinematics properties of mechanisms.

The simulation tools provide powerful solutions for planning and designing of complex assembly facilities, lines and workplaces. In the digital mockup context, the haptic interfaces can be used to reduce the development cycle of products.

The simulation tools have taken, for a few years, a great place in the design process of robotized environments. In order to define a manipulator task, in the CAO-Robotics applications, we have tools to place the end-effector of the robot, according to its joint co-ordinates (direct geometric model) or to know the manipulator postures, according to the localization of its end-effector (inverse geometric model).

These two tools are used (1) to make trajectory planning, (2) to make optimal placement of mechanisms while being coupled with collision detection tools and (3) to know the boundaries of the mechanism workspace. For this last case, it is not possible to represent them when the manipulator has more than 3 degrees of freedom or only in a degraded way by representing the dexterous workspace.

Other tools must be used in order to understand the behavior real of the manipulator. The detection of the singular configurations is an important problem because in this case, exist some directions, for which the robot cannot move or can become difficult to control (vibrations).

The problem of these tools (when it exists) is the representation and the interpretation of the results by the engineer. The important for the engineer is not only the value on the point but its variation and the information about the most favorable directions. The exploitation of the results is not obvious and the link with the performance value of the real phenomenon is not trivial.

When one analyzes the communication between the operator and the computer, he can perceive that the operator immersion in the digital model is very weak. In the majority of the cases, the operator only has a keyboard, a classic mouse or a 3D mouse to manage the displacements of the objects. For the visual side, the use of stereoscopic glasses adds the necessary three-dimensional vision to the apprehension of the distances (and therefore of proximity). The displacement of objects or manipulators in the space is a difficult problem when rotations and translations are coupled. Only the use of haptic device as a Phantom desktop allows a realistic displacement of these objects. These displacements can be limited by the collisions between the manipulator and its environment or between the segments of the manipulator. Several works have already been achieved in this sense, but they are limited to the reproduction of merely material phenomena [1].

In this paper, we show a paradigm how a haptic device with 3 or 6 degrees of freedom can allow the operator to interpret the limit of the workspace of the manipulators (including position and orientation). Then, we will add the interpretation of the manipulator mechanical conditioning while introducing some stimuli (viscosity, vibration…) in its displacements. When the conditioning is high or that one approach of the joint limits the operator will be able to feel, while displacing the manipulator's effector, the proximity of a singular configuration. Thus, we will be able to materialize virtual phenomena (theoretical performances) while using a peripheral of virtual reality and we enrich the digital mockup by additional semantics. Indeed, the indications of mechanical performances are often difficult to represent using traditional means. Thanks to the integration of the haptic feedback in the environment of simulation, the user will have an additional tool for the materialization of these properties. The integration of the peripheral haptic feedback in the environment of simulation eM-Workplace was the subject of a specific development for the Tecnomatix [1]. The work presented in [1] is related to the manipulation of objects, robots and models in a cluttered workspace. The work that we present in this paper is an extension of the results gotten.

## 2- Collision detection and haptic force-feedback generation

### 2.1- Collision detection

Desktop virtual prototyping can reduce development costs and cycle times, increase design flexibility and facilitate a more interactive, concurrent engineering process. However, trying to manipulate parts in these scenarios by visual feedback alone is generally very difficult. Providing the operator with "sense of touch" feedback going well beyond mere collision detection. Ideally, this sense of touch should include force and torque feedback in the same six degrees of freedom that the object would have in free space-x, y, z, roll, pitch and yaw.

In order to perform the operation of "collision detection", several tools can be find. For example, the Voxel PointShell (VPS) software is used by Sensable in GHOST software developer's kit [2-4]. There are several semi-automatic or automatics methods, for the detection and the avoidance of collision in generation of trajectory. These methods are not yet sufficiently mature for a complete integration and a respect of the real time constraint [5-7].

In our application, we use the collision detection integrated in eM-Workspace to avoid the implementation of new algorithm. This function gives us several data. The characteristic of our approach is strongly related to the direct use of the models available in marketed software. Virtual a reality application should not require a new programming of all the models already offered in this software. Several aspects can be exploited from these models, such that:

- CAD models (geometrical aspect): These models are generally relatively heavy and can indeed decrease the real times performances in a VR environment. On the other hand, the transfer of these models in software not dedicated leads to a lost of data and are very times consuming.

- Non-geometrical data relating to the kinematics and dynamics behavior of the objects, robots and mechanisms. The CAO-Robotics environments integrate and manage these data complementarily with the geometrical data.

- The behavior models are also a specificity of CAO-Robotics software (kinematic and dynamic models, workspace, collision detection...). The fact that this software can manage the majority of the industrial robots is important and provides a considerable advantage to integrate new VR functions inside.

Conversely, each model introduces additional constraints into the VR environment by increasing the times response of the user interface. Our choice is a necessary compromise, which preserves the entirety and the exactitude of the models and which exploits simplified or approximate functions (example collision detection).

When the user activates force feedback, he must select an environment of collisions. This popup appears when "Env." button has been clicked. It enables to choose the pair, which could be in collision with selected object. Here, the idea is that two objects are needed to have a collision. Then with selected object and environment, the object could be in collision with environment. The Toolbox/collisions command can check for collisions and near misses either between all of the components in the workcell, or between the items in the lists prepared by these commands. Checking collisions and near misses among prepared lists of components is more efficient, if the lists are small, than checking among every component in the workcell, since it reduces the computational overhead that would otherwise be required to check for collisions and near misses between irrelevant and uninteresting components.

### 2.2- Force-feedback

The user interface of virtual reality software can be divided into two main classes (i) the input and (ii) the output. In the context of CAD/CAM software, the number of these devices is limited. In the context of virtual reality applications, the haptic sense is a powerful tool to believe that something is "real". The visual display is the main device used in CAD/CAM software. However, for a virtual reality software, there exist five categories of visual display, (i) monitor-based VR, (ii) projector VR (stationary displays), (iii) occlusive HMDs, (iv) nonoclusive HMDs (head-based displays), and (v) palm VR (hand based displays).

There are three types of haptic displays, (i) tactile devices, (ii) end-effector displays, and (iii) robotically operated shape displays. The first one provides information to the user in response to touching, grasping, feeling surfaces, or sensing the temperature of an object. The second one provides resistance and pressure to achieve these effects. The last one uses robots to present physical object to the user fingertips and provides information about shape, texture, and location to the user.

The haptic device used provides 6 degree of freedom positional sensing but only tree degree force feedback and a small workspace. To avoid these problems, an intuitive strategy is used to lock the rotation of the stylus when an object is in the vicinity of environment and to manage the stylus/object attachment. The Phantom desktop workspace is (16 cm x 13 cm x 13 cm). The relative precision of position is 0.02 mm and the force-feedback is 6.4 N maxi and 1.4 N in continuous.

## 3- eM-Virtual Desktop application

The eM-Worplace provides powerful solutions for planning and designing of complex assembly facilities, lines and workplaces. In the digital mockup context, the haptic interfaces can be used to reduce the development cycle of products. In the literature, we can find several prototypes of software that integrate the haptic devices but, currently, not any software exist in the context of CAD/CAM software. Thus, to reduce the development cycle of products is necessary to limit the number of software to avoid the limitations of data exchanges.

Two haptic devices are implemented in eM-Virtual Desktop. The first one is the Phantom desktop device (Figure 1), a force-feedback haptic device of SensAble Technologies [3]. The second is the CyberGlove device (Figure 2) [4]. In this article, only the implementation of the first device is illustrated.

All the Phantom products, from the Sensable society [3] have the same architecture. The Phantom is grasped via a stylus on which extremity a switch is located. The device used in our application is the Phantom desktop. It is a six-dof input device since both the position and the orientation of the user's hand are sensed, but is a tree-dof force-feedback device, since only the forces along the x-, y- and z-axes are sent back to the user. Thus, it cannot mechanically return any torque information to the user. For the exploration of application areas requiring force feedback in six degrees of freedom (6DOF), there is phantom 1.5/6DOF but is not yet implemented in our software [6].

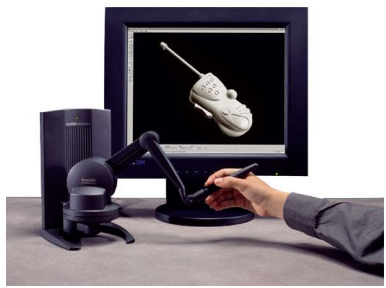 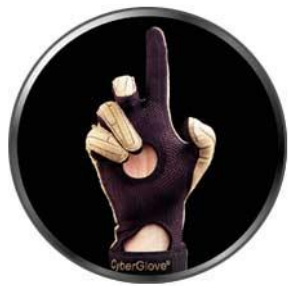

**Fig.1: Phantom desktop**          **Fig. 2: Cyberglove**

The software eM-Virtual desktop in completely integrated in eM-Workplace. It is an addition application that can be addressed on the list of available applications. Figure 3 depicts de communication protocol used for this application (see figure 3 and [1]).

The General Haptic Open Software Toolkit developed by SenAble (GHOST) includes several tools to make the collision detection in a simple environment. The implemented objects are simple shape like cylinder, sphere, or vrml geometry objects. In an industrial environment, such tool cannot be used due to the complexity of digital mockup and its object motions on the cell. As a matter of fact, several processes can carry out in the same time to manage an assembly cell for example. To solve this problem, we will use the collision algorithm detection of eM-Workplace. In fact, not any geometric entities are replicated and we work on a uniqueness database. The main problem of this architecture is the implementation of robust and fast interface between the GHOST and eM-Workplace, which is presented in the article. Three different loops are used to manage the graphics; the collision detection and the haptic feedback according to theirs own frequencies.

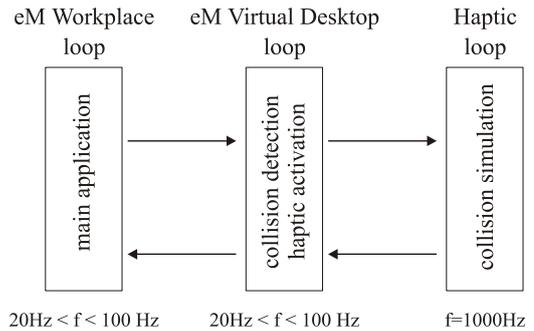

**Fig.3: Communication protocol**

There are three levels of <u>sensibility</u> (Rough, medium and fine) on the translation movement. It enables to choose the sensibility of the relation between the screen and the movement of Phantom desktop. The option "screen" is a special sensibility: Following the dimension of the view (zoom-/zoom+), the sensibility is automatically calculated proportionally with the zoom of the view.

So the user can define coefficients for all items (Rough, Medium, Fine).

### 3.1- Managing objects, robots and mannequin

Three types of object can be managed by the Phantom desktop via the motion of a stylus in a cell. For each type, a mechanical comportment is associated. The simplest is an object, which can move with respect to its geometric center or its self-origin or a frame defined by the operator (fig.4).

The second one is a robot, which can be moved via its end-effector (TCPF) in using the inverse kinematic model or its base when the robot is compared with a rigid body (fig. 5).

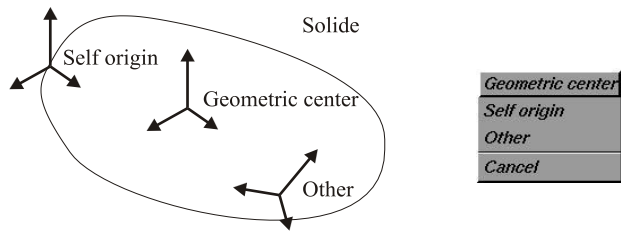

**Fig. 4: Reference frames associated to the solid**

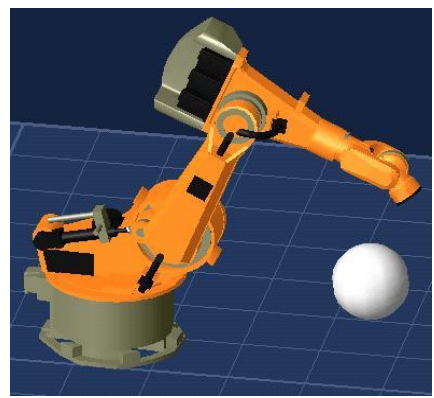

**Fig. 5: Robot in eM-Virtual Desktop**

The third one is a mannequin, which can manage either its left hand or right hand. The inverse kinematic model of the mannequin can managed its 56 degree of freedom with respect to ergonomic constraints and lock either one hand or the knees or the torso (Fig. 6).

The application eM-Human provides a virtual environment for interactive design and optimization of manual tasks. In a 3D model of the actual manufacturing environment, you can define the work sequence using a virtual human model. Comprehensive functions allow for accurate analysis of the workplace in regards to execution time and ergonomics of the human task. The impact of modifications can be checked instantly, thus enabling the planner to optimize the work system prior to implementation.

eM-Human provides a 3D virtual environment wherein you can design and optimize manual operations. A library of human models of different gender and sizes, based on international standards, ensures that the workplace design is suitable for a broad range of workers. The human models provide inverse kinematics and posture calculations for the complete body, enabling detailed, accurate and efficient design of human tasks. Different grasping and walking macros allow fast and simple definition of human motions.

eM-Human provides capabilities to detect collisions between the human and the environment, and to analyze reachability, ensuring the feasibility of human tasks. A separate window on the screen, showing the worker´s field-of-vision, allows close examination of tasks from the worker´s point of view.

With the application eM-Virtual Desktop, all the displacement of the manikin are available via the stylus of the Phantom desktop (*Fig. 6*).

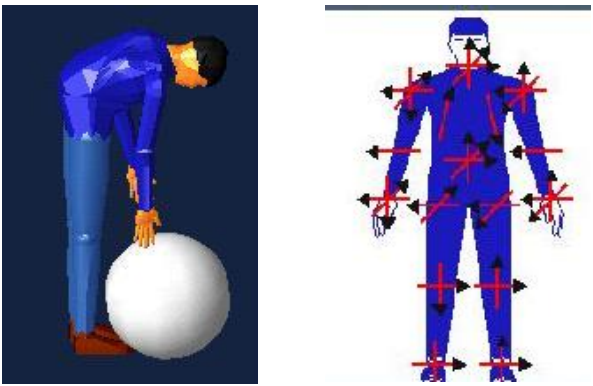

**Fig. 6: A mannequin in the eM-Virtual Desktop and its 56 degrees of freedom**

## 4- Mechanism properties

Haptic device in the environment of robotic simulation introduces a new approach to feel and to understand the boundaries of the workspace of mechanisms as well as its kinetostatic properties. In the following, the application of the haptic force-feedback device corresponding to these properties is presented.

### 4.1- Mechanism definitions

The mechanisms (or robots) are widely used in various industrial applications. Since last decade, other areas of application have emerged: medical, service, transport, underwater, entertainment… To make path planning, two methods are possible, on line or off line. The first stops the mechanism during its programming whereas the second brings more flexibility because it can be to realize on another site without stopping the production.

A mechanism consists of two distinct subsystems, one (or more) end-effector and an articulated mechanical structure. There exist various types of mechanism (serial, tree structured and closed chains), but a great part of its properties are common.

### 4.2- Direct and inverse geometric model and workspace

The control of a mechanism requires the computation of two mathematical models. The transformation models between the joint space and the task space. Indeed, these transformation models are very important since the mechanism is controlled in the joint space whereas tasks are defined in the task space.

Two classes of models are studied, (i) direct and inverse geometric model, and (ii) direct and inverse kinematic models. In the context of the simulation, it is often the first one, which is used.

To select or to program a mechanism, the workspace and its singularity branches must be studied.

Let $\mathbf{q} = \begin{bmatrix} q_1 & \cdots & q_n \end{bmatrix}$ be an element of the joint space and let $\mathbf{X} = \begin{bmatrix} x_1 & \ldots & x_m \end{bmatrix}$ is the corresponding element in the task space, with $\mathbf{X} = f(\mathbf{q})$.

The joint domain $\mathbf{Q}$ is define as the set of all reachable configuration taking into account the joints limits:

$$\mathbf{Q} = \{\mathbf{q} \mid q_{i\min} \leq q_i \leq q_{i\max}, \forall i = 1,...,n\}$$

The image of $\mathbf{Q}$ by the direct geometric model defines the workspace $\mathbf{W}$ of the robot, with

$$\mathbf{W} = f(\mathbf{Q}).$$

The workspace $\mathbf{W}$ is the set of the position and the orientation reachable by the end-effector. Its shape depends on the architecture and its boundary is defined by singularities and the joints limits. The singular branches are defined by calculated the determinant of the Jacobian matrix $\mathbf{J}$,

$$\mathbf{J} = \begin{bmatrix} \mathbf{e}_1 & \cdots & \mathbf{e}_n \\ \mathbf{e}_1 \times \mathbf{r}_1 & \cdots & \mathbf{e}_n \times \mathbf{r}_n \end{bmatrix}$$

where $\mathbf{e}_i$ are associated with each revolute of prismatic joints and $\mathbf{r}_i$ is defined as that joining $O_i$ with $P$, directed from the former to the latter (Fig. 7).

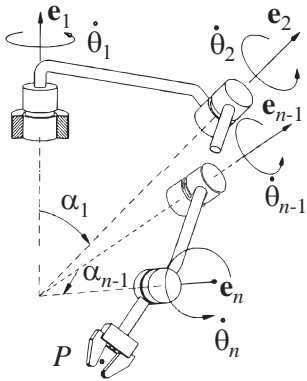

**Fig. 7: A general n-revolute spherical wrist**

However, when there is an obstacle in its workspace, additional boundaries limiting the reachable zones appear [8, 9]. In general case, the workspace of a robot is a 6-dimmensionnal space (position and orientation of the end-effector), which is difficult to handle. So usually, we must study its projection in the 3-dimmensional-position space (fig. 8). In our first study only the last type of boundary are introduced in the simulation software. However, to be complete, we need to introduce all the boundary types to feel all the properties of the mechanism.

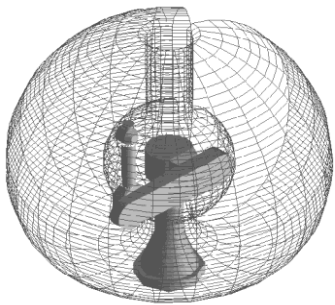

**Fig. 8: Workspace of a 6-dof mechanism**

*4.3- Kinetostatic indices*

Various performance indices have been devised to assess the kinetostatic performance of serial manipulators. Among these, the concepts of *service angle* [8], *dexterous workspace* [9] and *manipulability* [10] are worth mentioning. All these different concepts allow the definition of the kinetostatic performance of a manipulator from correspondingly different viewpoints. However, with the exception of Yoshikawa's manipulability index [10], none of these considers the invertibility of the Jacobian matrix. A dimensionless quality index was recently introduced by Lee [11] based on the ratio of the Jacobian determinant to its maximum absolute value, as applicable to parallel manipulators. This index does not take into account the location of the operation point in the end-effector, for the Jacobian determinant is independent of this location. The proof of the foregoing fact is available in [12], as pertaining to serial manipulators, its extension to their parallel counterparts being straightforward. The *condition number* of a given matrix is well known to provide a measure of invertibility of the matrix [13]. It is thus natural that this concept found its way in this context. Indeed, the condition number $\kappa(\mathbf{J})$ of the Jacobian matrix was proposed by Salisbury [14] as a figure of merit to minimize when designing manipulators for maximum accuracy,

$$\kappa(\mathbf{J}) = \frac{\sigma_l}{\sigma_s},$$

where $\sigma_l$ is the largest singular value of the Jacobian matrix $\mathbf{J}$ and $\sigma_s$ the smallest one.

In fact, the condition number gives, for a square matrix, a measure of the relative round off error amplification of the computed results [13] with respect to the data round off error. As is well known, however, the dimensional inhomogeneity of the entries of the Jacobian matrix prevents the straightforward application of the condition number as a measure of Jacobian invertibility. The *characteristic length*, noted $L$, was introduced in [15] to cope with the above-mentioned non-homogeneity.

$$\mathbf{J} = \begin{bmatrix} \mathbf{e}_1 & \cdots & \mathbf{e}_n \\ \frac{1}{L}\mathbf{e}_1 \times \mathbf{r}_1 & \cdots & \frac{1}{L}\mathbf{e}_n \times \mathbf{r}_n \end{bmatrix}$$

Apparently, nevertheless, this concept has found strong opposition within some circles, mainly because of the lack of a direct geometric interpretation of the concept. However, recent papers are introduced for serial and parallel mechanism, a novel performance index that lends itself to a straightforward manipulation and leads to sound geometric relations. Briefly stated, the performance index proposed here is based on the concept of *distance* in the space of $m \times n$ matrices, which is based, in turn, on the concept of inner product of this space. The performance index underlying this paper thus measures the distance of a given Jacobian matrix from an isotropic matrix of the same gestalt. With the purpose of rendering the Jacobian matrix dimensionally homogeneous, we resort to the concept of posture-dependent *conditioning length*. Thus, given an arbitrary serial manipulator in an arbitrary posture, it is possible to define a unique length that renders this matrix dimensionally homogeneous and of minimum distance to isotropy. The characteristic length of the manipulator is then defined as the conditioning length corresponding to the posture that renders the above-mentioned distance a minimum over all possible manipulator postures.

## 5- Introduction of the realistic rendering concept

The CAD-CAM software allows the simulation of mechanism through their direct and inverse geometric models. In the first case, it can exist of the dialog boxes with circular or linear sliders in which a change of color appears when one approaches these limits. In the second case, i.e. the use of the inverse geometrical model, we can represent only boundary spaces of dimensions 3.

### 5.1- Force-feedback corresponding to joint limits

When the mechanism reaches its joint limits or the workspace limits, two types of information can be given to the user: visual or resonant. However, no information allows him to find a solution to continue his displacement [9]. In the working space, it is difficult to feel the proximity of the boundaries limiting the reachable zones.

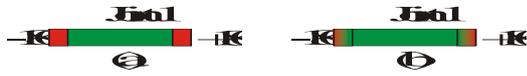

**Fig. 9: Joint limits representation
(a) with classic limits (b) with smooth limits**

This problem of joint limits of the manipulator is resolved in our application by making direct analogy with a physical collision. So when one of the links reaches its joint limit, the user must feel a force-feedback indicating him a collision. That, by sensing the force-feedback the user is not obliged to supervise all visual information on the screen. He immediately feels the joint limits.

In order to inform the user before arriving to the joint limit, we introduced a threshold security value close to the joint limit (Fig. 10).

Let us define: $q_{i\_max}$ and $q_{i\_min}$ two extreme joint limits corresponding to the joint $q_i$ and $\Delta_{qi}$ the threshold security value of $q_i$. The force-feedback felt by the user is null in the interval $[(q_{imin} + \Delta_{qi})\ (q_{imax} - \Delta_{qi})]$. This force increases linearly until a maximal value. Thus the user can adapt his strategy in order to generate the trajectory.

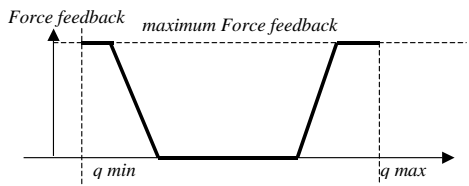

**Fig. 10: Smooth Force feedback for the joint limits**

### 5.2- Boundaries of the workspace and singular configurations

For uncoupled architectures or for architectures with three dof, one can visualize the limit of the prismatic workspace (for example the workspace of the robot Puma and the Orthoglide). In the case of uncoupled architecture (Diestro and platform of flight (fig. 11 and 12), one can only materialize their workspace corresponding to a given orientation.

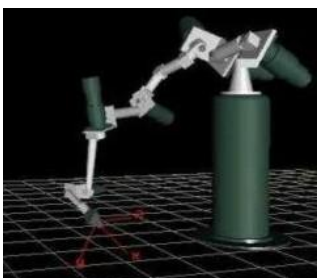

**Fig. 11: The Diestro robot**

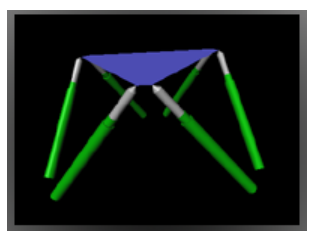

**Fig. 12: The Gough-Stewart platform**

From Figure 8 one can see that the boundaries of the workspace is not easy to analyse and to takes into account. One can notice that the complexity increase for robots with internal boundaries of the workspace and for Gough-Stewart. This workspace is also function of the orientation.

In our application, we consider this workspace as a physical obstacle and we provide the same tools, presented in section 2, in order to avoid these limits. The conditioning number is used to take into account sensibility analysis. To prevent the critical situations with "*bad*" condition number, a threshold security value of the conditioning number is introduced. The force-feedback value is deduced from the condition number. Its value is maximal when the condition number is greater than the threshold security value (fig. 13). Since:

$$\kappa(\mathbf{J}) = \frac{\sigma_l}{\sigma_s},$$

where $\sigma_l$ is the largest singular value of the Jacobian matrix **J** and $\sigma_s$ the smallest one then:

$$0 \leq 1/\kappa(\mathbf{J}) \leq 1$$

The limits of the workspace are obtained when $1/\kappa(\mathbf{J})$ is close to zero. The best performances are in the region where $1/\kappa$ is close to 1. One can also notice that if $1/\kappa(\mathbf{J})$ decrease very quickly, the robustness of the performances of the configuration in the corresponding situation is very sensible to the local displacements.

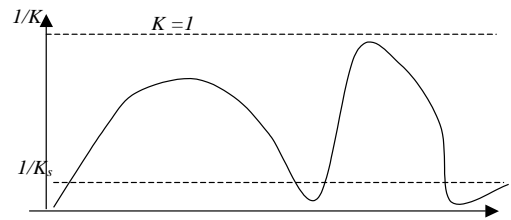

**Fig. 13: Force feedback corresponding to the conditioning number performance**

## 6- Application to a 2-dof parallel robot

Parallel kinematic machines (PKM) are commonly claimed to offer several advantages over their serial counterpart, like high structural rigidity, high dynamic capacities and high accuracy. On the other hand, they generally suffer from a reduced operational workspace due to the presence of internal singularities or self-collisions. The mechanism under study is a simple 2-dof parallel mechanism for which not any kinetostatic performance index is constant throughout its workspace. In next section, the kinematic equations are presented and the value of the kinetostatic performance index used, i.e. the condition number of two matrices.

### 6.1- Kinematic Relations

The manipulator under study is a five-bar, revolute (*R*)-coupled linkage, as displayed in Fig. 14. The actuated joint variables are $\theta_1$ and $\theta_2$, while the Cartesian variables are the $(x, y)$ coordinates of the revolute center $P$.

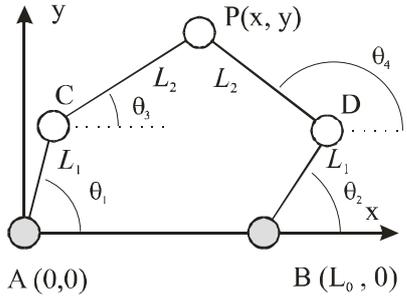

**Fig. 14: A two-dof closed-chain manipulator**

Lengths $L_0$, $L_1$, $L_2$, $L_3$, and $L_4$ define the geometry of this manipulator entirely. However, in this paper we focus on a symmetric manipulators, with $L_1 = L_3$ and $L_2 = L_4$. The symmetric architecture of the manipulator at hand is justified for general tasks. In manipulator design, then, one is interested in obtaining values of $L_0$, $L_1$, and $L_2$ that optimize a given objective function under some prescribed constraints.

The velocity $\dot{\mathbf{p}}$ of point $P$, of position vector $\mathbf{p}$, can be obtained in two different forms, depending on the direction in which the loop is traversed, namely,

$$\dot{\mathbf{p}} = \dot{\mathbf{c}} + \dot{\theta}_3 \mathbf{E}(\mathbf{p}-\mathbf{c}) \quad \text{and} \quad \dot{\mathbf{p}} = \dot{\mathbf{d}} + \dot{\theta}_4 \mathbf{E}(\mathbf{p}-\mathbf{d})$$

with matrix $\mathbf{E}$ defined as,

$$\mathbf{E} = \begin{bmatrix} 0 & -1 \\ 1 & 0 \end{bmatrix}$$

and $\mathbf{c}$ and $\mathbf{d}$ denoting the position vectors, in the frame indicated in Fig. 14, of points $C$ and $D$, respectively. After simplification, we write the kinematic equations in vector form, such that,

$$\mathbf{A}\dot{\mathbf{p}} = \mathbf{B}\dot{\boldsymbol{\theta}}$$

with $\dot{\boldsymbol{\theta}}$ defined as the vector of actuated joint rates, of components $\dot{\theta}_1$ and $\dot{\theta}_2$. Moreover $\mathbf{A}$ and $\mathbf{B}$ are, respectively, the direct-kinematics and the inverse-kinematics matrices of the manipulator, defined as,

$$\mathbf{A} = \begin{bmatrix} (\mathbf{p}-\mathbf{c})^T \\ (\mathbf{p}-\mathbf{d})^T \end{bmatrix} \text{ and } \mathbf{B} = L_1 L_2 \begin{bmatrix} \sin(\theta_3 - \theta_1) & 0 \\ 0 & \sin(\theta_4 - \theta_2) \end{bmatrix}$$

The singular configurations associated to $\mathbf{A}$ (resp. $\mathbf{B}$) are called direct-kinematic singularities (resp. inverse-kinematic singularities).

The direct-kinematic singularities are located inside the Cartesian workspace and the inverse-kinematic singularities are often located on its boundaries. In a direct-kinematic singularity, the mobile platform wins one or several degrees of freedom and thus become incontrollable. On such configurations, the forces inside the mechanism tend to infinity. However, this kind of behavior is not trivial to detect without a VR interface.

In the next sections, the condition number of both matrices is derived to provide us specific indices to render these boundaries. These indices are used to generate the value of force feedback to define a user friendly interface.

### 6.2- Direct-Kinematics Matrix

To calculate the condition number of matrix $\mathbf{A}$, we need the product $\mathbf{A}\mathbf{A}^T$, which we calculate below:

$$\mathbf{A}\mathbf{A}^T = L_2^2 \begin{bmatrix} 1 & \cos(\theta_3 - \theta_4) \\ \cos(\theta_3 - \theta_4) & 1 \end{bmatrix}$$

The eigenvalues $\alpha_1$ and $\alpha_2$ of the above product are given by:

$$\alpha_1 = 1 - \cos(\theta_3 - \theta_4) \text{ and } \alpha_2 = 1 + \cos(\theta_3 - \theta_4)$$

and hence, the condition number of matrix $\mathbf{A}$ is,

$$\kappa(\mathbf{A}) = \sqrt{\frac{\alpha_{max}}{\alpha_{min}}} = \sqrt{\frac{1}{\tan\left(\frac{\theta_3 - \theta_4}{2}\right)}}$$

where $\alpha_{min} = 1 - |\cos(\theta_3 - \theta_4)|$ and $\alpha_{max} = 1 + |\cos(\theta_3 - \theta_4)|$.

In light of above expression for the condition number of the Jacobian matrix $\mathbf{A}$, it is apparent that $\kappa(\mathbf{A})$ attains its minimum of 1 when $|\theta_3 - \theta_4| = \pi/2$, the equality being understood *modulo* $\pi$. At the other end of the spectrum, $\kappa(\mathbf{A})$ tends to infinity when $\theta_3 - \theta_4 = k\pi$, for $k = 1, 2$. When matrix $\mathbf{A}$ attains a condition number of unity, it is termed *isotropic*, its inversion being performed without any round-off-error amplification. Manipulator postures for which condition $|\theta_3 - \theta_4| = \pi/2$ holds are thus the most accurate for purposes of the direct kinematics of the manipulator. Correspondingly, the locus of points whereby matrix $\mathbf{A}$ is isotropic is called the *isotropy locus* in the Cartesian workspace. On the other hand, manipulator postures whereby $\theta_3 - \theta_4 = k\pi$ denote a singular matrix $\mathbf{A}$. Such singularities occur at the boundary of the Joint space of the manipulator, and hence, the locus of $P$ whereby these singularities occur, namely, the *singularity locus* in the Joint space, defines this boundary. Interestingly, isotropy can be obtained regardless of the dimensions of the manipulator, as long as (i) it is symmetric and (ii) $L_2 \neq 0$.

### 6.3- Inverse-Kinematics Matrix

By virtue of the diagonal form of matrix $\mathbf{B}$, its singular values, $\beta_1$ and $\beta_2$, are simply the absolute values of its diagonal entries, namely,

$$\beta_1 = |\sin(\theta_3 - \theta_1)| \text{ and } \beta_2 = |\sin(\theta_4 - \theta_2)|$$

The condition number $\kappa$ of $\mathbf{B}$ is thus

$$\kappa(\mathbf{B}) = \sqrt{\frac{\beta_{max}}{\beta_{min}}}$$

where, if $|\sin(\theta_3 - \theta_1)| < |\sin(\theta_4 - \theta_2)|$, then

$$\beta_{min} = |\sin(\theta_3 - \theta_1)| \text{ and } \beta_{max} = |\sin(\theta_4 - \theta_2)|$$

else

$$\beta_{min} = |\sin(\theta_4 - \theta_2)| \text{ and } \beta_{max} = |\sin(\theta_3 - \theta_1)|.$$

In light of above expression for the condition number of the Jacobian matrix $\mathbf{B}$, it is apparent that $\kappa(\mathbf{B})$ attains its minimum of 1 when $|\theta_3 - \theta_1| = |\theta_4 - \theta_2| = 0$. The locus of points where $\kappa(\mathbf{B}) = 1$, and hence, where $\mathbf{B}$ is isotropic, is called the *isotropy locus* of the manipulator in the joint space. At the other end of the spectrum, $\kappa(\mathbf{B})$ tends to infinity when $|\theta_3 - \theta_1| = k\pi$ or $|\theta_4 - \theta_2| = k\pi$, for $k = 1, 2$, which denote singularities of **B**. These singularities are associated with the inverse kinematics of the manipulator, and hence, lie within its Cartesian workspace, not at the boundary of this one. The singularity locus of **B** thus defines the Cartesian workspace of the manipulator. Therefore, the Cartesian workspace of the manipulator is bounded by the singularity locus of **B**, i.e., the locus of points where $\kappa(\mathbf{B}) \to \infty$. Interestingly, **B** can be rendered isotropic regardless of the dimensions of the manipulator, as long as (i) it is symmetric and (ii) $L_1 \neq 0$ and $L_2 \neq 0$.

*6.4- Example with the direct kinematic matrix*

We assume here the dimensions $L_0 = 6$, $L_1 = 8$, and $L_2 = 5$, in certain units of length that we need not specify. The iso-conditioning curves for the direct-kinematic matrix both in the Cartesian and in the joint spaces are displayed in Figs. 15.

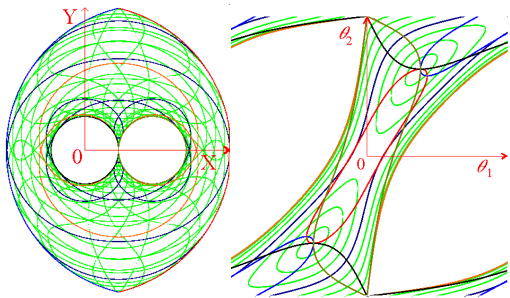

**Fig. 15: The iso-conditioning curves in the Cartesian and joint space**

A better representation of iso-conditioning curves can be obtained in the Cartesian space by displaying these curves for every working mode [16], which we do in Fig. 16.

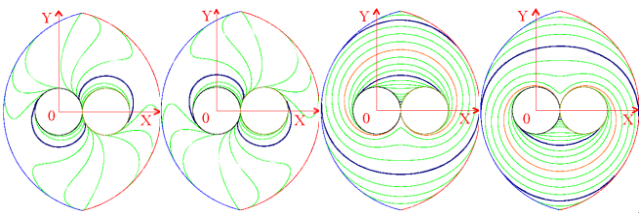

**Fig. 16: The four working modes and their iso-conditioning curves in the Cartesian space**

According to the working mode chosen, the behavior of the mechanism studied change.

## 7- Conclusions

We have shown in this study that some complex concept can be simplified in order to be popularized. Furthermore, in a general case, the visualization of complex concept is not possible. Thus, the utilization of user interface seems to be the most appropriated solution for this problem.

We have introduced in this paper, new tool for the engineer to fell and to understand the properties of mechanisms. Such tolls can be used in the industry as well as in the engineering school to illustrate some theoretical notions like singular configuration or condition number.

## 8- References